\documentclass[sigconf]{acmart}
\AtBeginDocument{%
  \providecommand\BibTeX{{%
    \normalfont B\kern-0.5em{\scshape i\kern-0.25em b}\kern-0.8em\TeX}}}

\copyrightyear{2023}
\acmYear{2023}
\setcopyright{rightsretained}\acmConference[COLIEE 2023]{COLIEE 2023 workshop: Competition on Legal Information Extraction/Entailment}{June 19, 2023}{Braga, Portugal}
\acmBooktitle{Proceedings of COLIEE 2023 workshop, June 19, 2023, Braga, Portugal}
\acmPrice{}
\acmISBN{}
\acmDOI{}

\usepackage[normalem]{ulem}
\usepackage{CJKutf8} 
\usepackage{multirow}
\usepackage{svg}
\useunder{\uline}{\ul}{}
\usepackage{arydshln}

\begin{document}

\title{CAPTAIN at COLIEE 2023: Efficient Methods for Legal Information Retrieval and Entailment Tasks}


\author{Chau Nguyen} \authornote{Authors contributed equally to the paper}
\author{Phuong Nguyen} \authornotemark[1]
\author{Thanh Tran}\authornotemark[1]
\author{Dat Nguyen}\authornotemark[1]
\author{An Trieu}\authornotemark[1]
\author{Tin Pham}
\author{Anh Dang}
\author{Le-Minh Nguyen}\authornote{The corresponding author}
\affiliation{%
  \texttt{\{chau.nguyen,phuongnm,thanhtc,nguyendt,antrieu,tinpham,hoanganhdang,nguyenml\}@jaist.ac.jp}\\
  \institution{Japan Advanced Institute of Science and Technology}
  \city{Ishikawa}
  \country{Japan}
}

\renewcommand{\shortauthors}{CAPTAIN Team}

\begin{abstract}
The Competition on Legal Information Extraction/Entailment (COLIEE) is held annually to encourage advancements in the automatic processing of legal texts. Processing legal documents is challenging due to the intricate structure and meaning of legal language. In this paper, we outline our strategies for tackling Task 2, Task 3, and Task 4 in the COLIEE 2023 competition. Our approach involved utilizing appropriate state-of-the-art deep learning methods, designing methods based on domain characteristics observation, and applying meticulous engineering practices and methodologies to the competition. As a result, our performance in these tasks has been outstanding, with first places in Task 2 and Task 3, and promising results in Task 4. 
Our source code is available at \url{https://github.com/Nguyen2015/CAPTAIN-COLIEE2023/tree/coliee2023}.
\end{abstract}





    \maketitle

\section{Introduction}
The COLIEE competition \cite{kim2023coliee} (Competition on Legal Information Extraction/Entailment) is an annual event that focuses on the automated processing of legal texts. The competition involves two types of data: case law and statute law. For each type of data, there are two tasks: legal information retrieval and legal information entailment. 

Task 1 involves retrieving cases that support a given case law, which is a crucial aspect of legal practice, used by attorneys and courts in decision-making. Task 2 also uses case law data, but requires models to find paragraphs that entail the decision of a given case. Tasks 3 and 4 involve using statute law data and tackling challenges such as retrieval and entailment. It is worth noting that Task 1 can be considered as an initial step towards Task 2, and Task 3 can serve as a preliminary stage for Task 4. Nonetheless, it is not obligatory to perform Task 1 before Task 2 or to undertake Task 3 prior to Task 4.

This year, we participated in Task 2, Task 3, and Task 4. The CAPTAIN team has succeeded in achieving competitive results in the COLIEE 2023 competition by utilizing appropriate deep learning techniques, suitable methods based on domain characteristics, and robust engineering practices and methodologies.

In Task 2, to tackle the challenging task of legal case entailment in COLIEE 2023, we propose an approach based on the pre-trained MonoT5 sequence-to-sequence model, which is fine-tuned with hard negative mining and ensembling techniques. The approach utilizes a straightforward input template to represent the point-wise classification aspect of the model and captures the relevance score of candidate paragraphs using the probability of the "true" token (versus the "false" token). The ensembling stage involves hyperparameter searching to find the optimal weight for each checkpoint. The approach achieves state-of-the-art performance in Task 2 this year, demonstrating the effectiveness of the proposed techniques.

Our solution for Task 3 addresses the issue of \textit{data diversity} across many legal categories (e.g., rights of retention, ownership, juridical persons). We observed that the queries and articles in the Civil Code include multiple categories, and the limited annotated data and lengthy pairs of questions and articles make it challenging for deep learning models to identify general patterns, often leading to local optima. To overcome this issue, we focused on constructing sub-models to learn specific aspects of the legal domain and designing an ensemble method to combine these sub-models to create a generalized system. We assumed that each local optimum of the legal retrieval system is biased towards particular categories, whereas our goal is to develop a system that covers all categories and utilizes the strengths of all sub-models through ensemble methods.

In Task 4, we experimented with three approaches, which involve identifying the entailment relationship between legal articles and a query. The first approach is to fine-tune language models with a novel augmentation mechanism. The second approach is based on condition-statement extraction, which involves breaking down legal articles and the query into components, identifying the entailing relationship based on the matching between the condition part and the statement part of the article and the query. The third approach is a hybrid approach that uses SVM ensembled with the condition-statement extraction method.

The CAPTAIN team proposed deep learning methods for these tasks and conducted detailed experiments to evaluate their approaches. The proposed methods could serve as a useful reference for researchers and engineers in the field of automated legal document processing.

\section{Related Works}
\subsection{Case Law} 
In COLIEE 2020, the Transformer model and its modified versions were widely used. TLIR \cite{shao2020thuir} and JNLP \cite{nguyen2020jnlp} teams used them for Task 1 to classify candidate cases with two labels (support/non-support). Team cyber \cite{westermann2020coliee} encoded candidate cases and base cases in TfIdf space and used SVM to classify, which helped them obtain the top rank in Task 1. For Task 2, JNLP \cite{nguyen2020jnlp} utilized the same approach as in Task 1 for the weakly-labeled dataset, which resulted in outperforming team cyber \cite{westermann2020coliee} and winning Task 2. 

In COLIEE 2021, teams used various BERT-based IR models to tackle Task 1, which is a case retrieval problem. TLIR team \cite{ma2021retrieving} achieved first place by applying an ensemble of multiple BERT models and selecting the highest result among them. Other teams also utilized BERT models and additional techniques such as text enrichment and similarity measures like Word Mover's Distance to improve their performance. In Task 2 of COLIEE 2021, the winning team NM \cite{rosa2021tune} utilized several models including monoT5-zero-shot, monoT5, and DeBERTa, and evaluated an ensemble of monoT5 and DeBERTa models. TR team \cite{schilder2021pentapus} used hand-crafted similarity features and a classical random forest classifier, while UA team \cite{kim2021bm25} used BERT pre-trained on a large dataset and a language detection model to remove French text. Siat team \cite{li2021siat} proposed a pre-training task on BERT with dynamic N-gram masking to obtain a BERT model with legal knowledge.

In COLIEE 2022, in Task 1, the winning team (UA \cite{rabelo2023semantic}) used a Transformer-based model to generate paragraph embeddings, then calculated the similarity between paragraphs of the query and the positive and negative cases. They used a Gradient Boosting classifier to determine if those cases should be noticed or not. The other teams used a combination of traditional and neural-based techniques, such as document and passage-level retrieval, adding external domain knowledge by extracting statute fragments mentioned in the cases, and re-ranking based on different statistical and embedding models. In Task 2, the winning team (NM \cite{rosa2022billions}) used monoT5, which is an adaptation of the T5 model. During inference, monoT5 generates a score that measures the relevance of a document to a query by applying a softmax function to the logits of the tokens “true” and “false”. They also extended a zero-shot approach and fine-tuned T5-base and T5-3B models for legal question-answering. The other teams used a combination of LegalBERT, BM25, and knowledge representation techniques such as Abstract Meaning Representation (AMR) to capture the most important words in the query and corresponding candidate paragraphs.

\subsection{Statute Law}
The approaches used in Task 3 of COLIEE 2020 involved various methods such as TfIdf, BM25, BERT models, and different word embedding techniques. The teams used different combinations of these techniques to classify articles as relevant or not based on the given legal queries. The LLNTU team \cite{shao2021bert} achieved the highest performance using the BERT model for article classification. The approaches used in Task 4 involved various NLP techniques and models such as BERT \cite{devlin2018bert}, RoBERTa \cite{liu2019roberta}, GloVe \cite{pennington2014glove}, and LSTM \cite{hochreiter1997long}. The winning team, JNLP \cite{nguyen2020jnlp}, employed BERT-based models fine-tuned with Japanese legal data and TfIdf to achieve the best performance. Other approaches also used rule-based ensembles, SVM \cite{cortes1995support}, and attention mechanisms with word embeddings to tackle the legal text classification task.

In Task 3 of the COLIEE 2021 competition, The OvGU team \cite{wehnert2021legal} took first place by using a variety of BERT models and data enrichment techniques, with the best run using Sentence-BERT \cite{reimers-2019-sentence-bert} embedding and TfIdf. The HUKB \cite {yoshioka2021hukb} system uses a BERT-based IR system with Indri \cite{strohman2005indri} and constructs a new article database. JNLP team \cite{nguyen2021jnlp} uses multiple BERT models and an ensemble approach for generating results. TR team \cite{schilder2021pentapus} uses Word Mover's Distance to calculate similarity and UA team \cite{kim2021bm25} uses ordinary IR modules, with the best run using BM25. In Task 4, the winning approach by HUKB \cite {yoshioka2021hukb} utilized an ensemble of BERT models with data augmentation. JNLP team \cite{nguyen2021paralaw} used a bert-base-japanese-whole-word-masking model with TfIdf-based data augmentation and proposed Next Foreign Sentence Prediction (NFSP) and Neighbor Multilingual Sentence Prediction (NMSP) for Task 4. OvGU \cite{wehnert2021legal} employed an ensemble of graph neural networks, where each node represented a query or article, and TR used existing models, including T5-based \cite{raffel2020exploring} ensemble, Multee \cite{trivedi2019repurposing}, and Electra \cite{clark2020electra}. UA team \cite{kim2021bm25} utilized BERT with semantic information. KIS team \cite{fujita2021predicate} extended their previous work using predicate-argument structure analysis, legal dictionary, negation detection, and an ensemble of modules for explainability.

In the COLIEE 2022 Task 3, the top-performing teams used various methods for answering legal questions. HUKB \cite{yoshioka2023hukb} and OvGU \cite{wehnert2023using} utilized information retrieval (IR) models with different variations in terms of document databases and sentence embeddings. JNLP \cite{bui2022using} proposed two separate deep learning models for answering ordinal questions and use-case questions, while LLNTU \cite{lin2022rethink} used a BERT-based approach with no clear adjustments for the current year's task. Finally, UA \cite{rabelo2023semantic} relied on IR models, specifically TfIdf and BM25. In Task 4, HUKB \cite{yoshioka2023hukb} proposed a method for selecting relevant parts from articles and employed an ensemble of BERT with data augmentation. JNLP \cite{bui2022using} compared different pre-trained models and data augmentation techniques. KIS \cite{fujita2023legal} employed an ensemble of rule-based and BERT-based methods with data augmentation and person name inference. LLNTU \cite{lin2022rethink} used the longest uncommon subsequence similarity comparison model, while OvGU \cite{wehnert2023using} employed graph neural networks on both the queries and articles, and additionally incorporating textbook knowledge.

\section{Methods\label{sec:method}}
\subsection{Task 2. Case Law Entailment Task}
The second task of COLIEE 2023 is legal case entailment: Given a fragment of a base case and a set of potential judicial-decision-related paragraphs of past cases, determine which candidate paragraphs entail the fragment of the base case. The dataset for this year's competition consists of 625 training cases with entailment labels and 100 test cases. The training set has an average ratio of 35-to-1 candidates per case and 1 entailment paragraph per case. As the ratio of candidates per case is high and the size of the dataset is limited, the semantic and structural complexity of legal documents makes this a particularly challenging task.

In order to overcome the challenge of this task, we utilize the pre-trained MonoT5 \cite{nogueira-etal-2020-document} sequence-to-sequence models on the MS MARCO \cite{nguyen2016ms} passage ranking dataset. MonoT5 is a novel approach to document ranking by fine-tuning the pre-trained T5 models with modified training data for the point-wise classification task, which outperformed the traditional BERT re-ranker approach. The process to convert a sequence-to-sequence T5 model to a point-wise classification MonoT5 model is straightforward with the following input template:

    \hfil "Query: [CASE] Document: [CANDIDATE] Relevant:"

The point-wise aspect of the model is represented by taking the probability of 2 tokens "true" and "false" of the decoded outputs. The relevance score of the candidate to the fragment of the base case is captured by the probability of the "true" token after normalization by the softmax function. After obtaining the relevance score of each candidate, the scores are sorted to form the final ranking list.

\begin{figure}[htbp]
  \centering
  \includegraphics[width=\columnwidth]{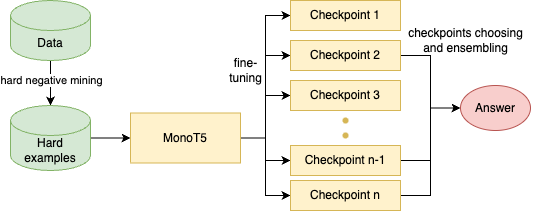}
  \caption{Our approach for Task 2}
  \label{fig:task2_method}
\end{figure}

Although MonoT5 has been used in past competitions \cite{coliee_2022_NM}, our approach introduces new engineering techniques in order to improve the performance of pre-trained checkpoints on legal entailment tasks, which we describe in the following sections. Figure \ref{fig:task2_method} provides an overview of our approach for Task 2.

\subsubsection{Fine-tuning with hard negative mining}
To form the training dataset for fine-tuning, we split the original dataset into training/validation/test segments and use the validation set to select the best checkpoints for later stages. We select the pre-trained MonoT5-large checkpoint on MS MARCO \cite{nguyen2016ms} as our initial weight for fine-tuning. Note that the pre-trained MonoT5-3B checkpoints are more powerful and outperformed the MonoT5-large variant on MS MARCO but unfortunately, we are not able to fine-tune these weights due to their high computation cost. We initialize the MonoT5-large model with the pre-trained weights and perform hard negative mining as follows. Firstly, we sample the top 10 negative examples of each base case's fragment (i.e., the top 10 fragments with a negative label that are most similar to the base case's fragment), sorted by the retrieval scores with BM25 to form the first training set and perform fine-tuning. After that, we re-sample the top 10 negative examples from the fine-tuned model from the previous step to form the final training set and fine-tune the original pre-trained weights with this dataset. This procedure has the purpose of mining the hard examples to further push the performance of the pre-trained models as these models have already achieved good performance in past competitions. After fine-tuning, we keep the 5 best checkpoints ordered by the metric score on the validation set for the next ensembling pipeline.

\subsubsection{Ensembling}
We perform ensembling the best checkpoints with hyperparameter searching on the validation set to get the ensembled score for each candidate. Each checkpoint is given an initial weight range from [0, 1] and the final weight is calculated by normalizing the initial weight with the sum of all initial weights. We use grid search to find the optimal weight for each checkpoint and save the ensembled scores for the final stage.

\subsubsection{Prediction}
After obtaining the candidate's scores for each base case, we sort the candidate list by score and select the top \emph{k} candidates. We keep the candidate with the highest score and for each remaining top candidate, we include the candidate in the final prediction if the difference between their relevance score to the highest score is smaller than a pre-defined margin \emph{m}. We perform a grid search on the values of \emph{k} and \emph{m} on the validation set to select the best values.

\subsection{Task 3. The Statute Law Retrieval Task\label{subs:3}} 

\paragraph{Problem statement} The objective of Task 3 in this competition is to extract a subset of Japanese Civil Code Articles that are relevant to a given legal bar exam question, denoted by Q. Participants are required to select articles, denoted by A$_i$ ($1 \leq i \leq n$), that are related to the given question Q.  This task serves as a precursor to Task 4, which requires participants to infer the legality of Q based on the relevant legal articles.

\subsubsection{Approach overview\label{subs:general_architecture}}  Following the previous approach \cite{nguyen2020jnlp,nguyen2021jnlp}, we also utilize the power of the pre-trained language model and some negative sampling techniques to fine-tune a model which can retrieve the relevant articles from a legal corpus. In detail, given an input query, top $k$ relevant articles (e.g., $k=150$ following \cite{nguyen2021jnlp})  are chosen based on term matching (e.g., TfIdf or BM25). Then, these articles are paired with the query and fed to a pre-trained language model to find the relevant articles.

\begin{figure*}[htbp]
  \centering
  \includegraphics[width=0.64\textwidth]{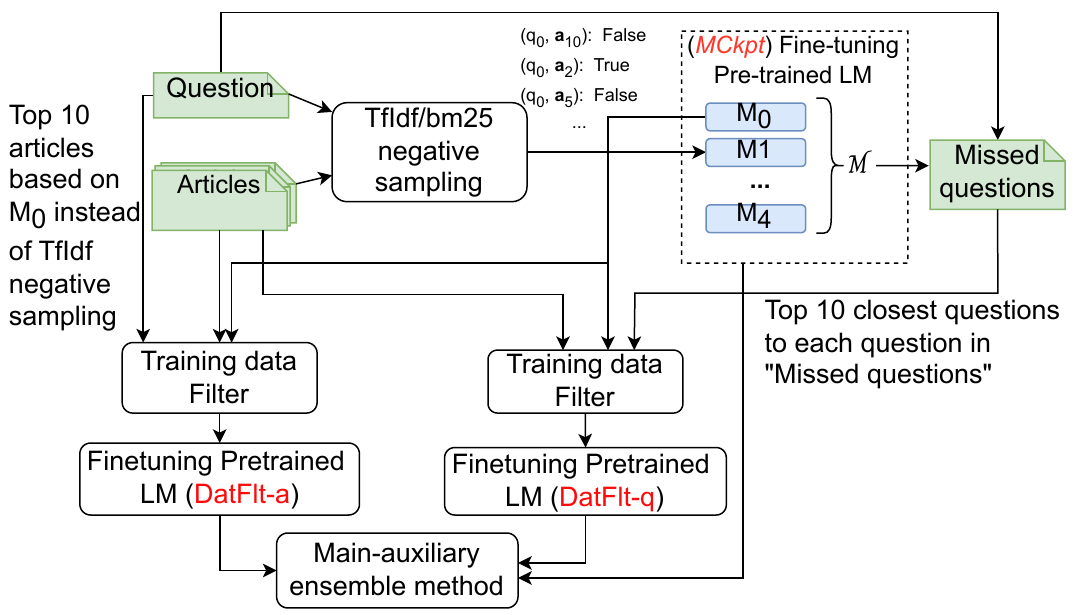}
  \caption{Our approach for Task 3\label{fig:proposedmethod}}
\end{figure*}

% PROBLEM and SOLUTION
\subsubsection{Approach details} 
\begin{figure}[htbp]
  \centering
  \includegraphics[width=0.45\textwidth,trim={4.8cm 21.3cm 4.6cm 4.cm},clip=true]{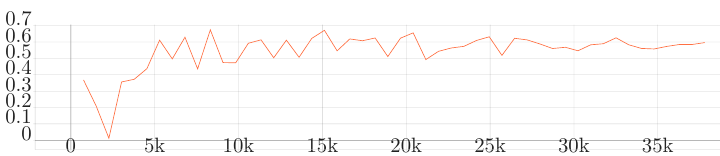}
  \caption{ Model performance (macro F2) on the development set with five epochs (about 35K steps).\label{fig:valid_f2}}
\end{figure}

As mentioned before, this year, our solution is focused on the problem of \textit{data diversity} in many categories. Based on our observation, we found that the queries and articles in the Civil Code include many categories. 
% Besides, the limited annotated data and the extremely long sample  (pairs of questions and articles) are also the reasons that make the deep learning model hard to find the generalized pattern and typically fall in the local optimums.  
In Figure~\ref{fig:valid_f2}, we show the different performance between checkpoints in the training process (which are local optimums). This result showed a large margin among the checkpoint's performance.

Therefore, in this work, we also introduce a simple yet effective method to ensemble model checkpoints in local optima to make a generalized system finally. We assume that each local optimum (fine-tuned model found in section~\ref{subs:general_architecture}) is biased to some categories, while our target is to build a system that can explore all the categories via sub-models and aggregate the strengths of these sub-models. To this end, we introduce some approaches to construct the sub-models to learn different aspects and design an effective way to combine them (Figure~\ref{fig:proposedmethod}). 

\paragraph{Model checkpoints (MCkpt)} As we mentioned, we assume that each checkpoint in the training process (Figure~\ref{fig:valid_f2}) is biased to some categories. Therefore, we combine the top $h$ best model checkpoints ($\mathcal{M} = \{M_i|0\leq i < h\}$) to collect these best states of a pre-trained language model based on the development set.
\paragraph{Pre-trained language models (Plm)} We hypothesize that using different pre-trained language models also supports the overall system to cover different aspects of data. Therefore, we design the combination method using two pre-trained language models \textit{cl-tohoku/bert-base-japanese-whole-word-masking}  and \textit{monot5-large-msmarco-10k}. 

\paragraph{Data filter (DatFlt)} In this approach, we aim to let the model learn within different data distributions to strengthen the robustness of the sub-model. To this end, we keep the original validation set and construct a new training set by two following strategies:
\begin{itemize}
    \item \textit{Tackling missed \textbf{q}uery (DatFlt-q)}: By using the fine-tuned mechanism described in section~\ref{subs:general_architecture}, we cached top $h$ model checkpoints in the training process (setting \textit{MCkpt} in Figure~\ref{fig:proposedmethod}, $\mathcal{M} = \{M_{i| 0\leq i < h}\}$). We found a set of questions that are typically \textit{``missed''} by these model checkpoints (i.e., the questions in which the fine-tuned models, $M_i$, did not find any relevant articles in the legal corpus). For each missed question, we filter the top 10  nearest questions to them in the training set (using cosine similarity of BERT \texttt{[CLS]} vector) to construct a new training dataset.  In particular, the new training set related to these questions is generated and we use it to continue to train with the best model checkpoint ($M_0$). To this end, we expect that the new training set has a similar topic/category to the \textit{missed} queries, which makes the model $M_0$ learn different features.
    \item \textit{Highly relevant \textbf{a}rticles in negative sampling (DatFlt-a)}: In difference with the previous approach,  this approach constructs a new training data set by filtering the highly relevant articles based on a fine-tuned model (e.g., top 10 nearest articles based on $M_0$) for the negative sampling step instead of the TfIdf scores. We expect this bootstrapping mechanism to boost the model and make it easier to separate similar articles in the legal corpus. 
\end{itemize}

\paragraph{Main-auxiliary ensemble method} Based on our experiments, we found that the overall system's performance will be hurt if we unite all the articles of all sub-models found in the previous step (because this standard combination method decreases the precision for each question sample). Therefore, we choose one sub-model (best performance based on the development set, $M_0$) as the main model, and we consider other sub-models as auxiliary models. Then we filter \textit{missed} questions of this model and union corresponding relevant articles of these questions from auxiliary models for the final result.

\subsection{Task 4. The Legal Textual Entailment Task}

We propose three approaches to tackle Task 4 of the COLIEE competition, which focuses on legal textual entailment. The objective of this task is to identify the entailment relationship between legal articles and a given query, where the answer is binary: "YES" (if the article entails the query) or "NO" (if the article does not entail the query). 

\subsubsection{First approach: Online data augmentation} In the previous work \cite{nguyen2020jnlp}, the authors proposed a mechanism to augment training data and fine-tune a pre-trained model in the legal domain. In line with this work, we introduce a mechanism named \textit{online data augmentation}, which re-samples the training data based on a pre-trained masked-language model (e.g., \textit{cl-tohoku/bert-base-japanese-whole-word-masking}). We argue that the main challenge in this task is the sparse annotated data distribution which is hard for the fine-tuning process. While in Task 3, each question can consider up to 100 or 150 relevant articles for negative samples, in this task, the number of relevant articles is usually only one or two. Therefore, the augmentation data approach is typically an appropriate solution to create a generalized system.

In terms of our data augmentation mechanism, given a question input, we used a pre-trained language model with the task recovery masked words \cite{devlin2018bert} to generate similar questions. In particular, in the training process, we randomly masked some words in a question and replaced them with a random selection from the top $k$ highest probabilities of candidate words output from a pre-trained language model. The generated question is then paired with the summaries of given relevant articles, and the generated pair is used for the fine-tuning process.

\subsubsection{Second approach: Condition-statement extraction}

\begin{figure}[htbp]
  \centering
  \includegraphics[width=\columnwidth]{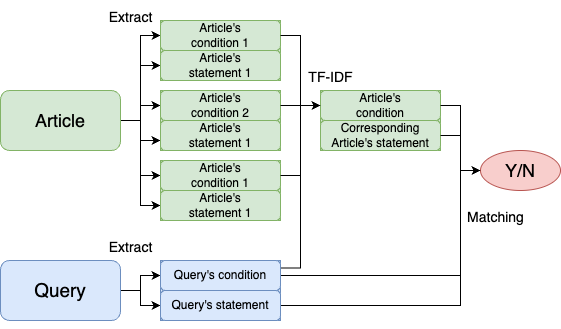}
  \caption{The condition-statement approach for Task 4}
  \label{fig:condition_statement}
\end{figure}

The method relies on the notion that a legal article can consist of multiple sentences, each of which discusses a particular statement subject to certain conditions. Additionally, a statement may have an exception if certain other conditions are met. An instance of this would be the phrase following "provided, however" in Table \ref{table_cond_state_example}, which represents the "other conditions" that contradict the primary statement "the person's residence is deemed to be the person's domicile".
In particular, we break down a legal article into multiple components, where each component consists of two parts: the condition part and the statement part. Similarly, we break down the query into two parts (condition and statement). Our approach involves determining the entailing relationship based on the "matching" (or entailing) between the condition part of the article and the query, as well as the statement part of the article and the query. We use a Semantic Role Labeling (SRL) model and heuristics to identify the relevant components in the legal articles and the queries, and then we employ a matching algorithm to determine the entailment relationship. Our approach shares similarities with the methods proposed in \cite{nguyen2018recurrent} and \cite{kim2019statute}. Nevertheless, our approach, which employs SRL to detect condition-statement pairs, has the potential to provide accurate and efficient results, especially when dealing with complex legal texts, which often contain lengthy and intricate sentences and structures. The overview of the approach is described in Figure \ref{fig:condition_statement}.

\textbf{Extract (condition, statement) pairs from an article.} The approach involves extracting (condition, statement) pairs from each point in an article, which usually consists of one sentence. To achieve this, the article is first split into sentences using the nltk library\footnote{https://github.com/nltk/nltk}. If a sentence does not have an exception, a Semantic Role Labeling technique is applied to identify the main verb, its arguments, and other components. Two heuristics are used: (i) the main verb is identified as the verb with the SRL labeling that covers most of the words in the input sentence, and (ii) the statement is extracted from the sentence where the first token is the first token of the main verb or argument, and the last token is the last token of the main verb or argument. The condition is then constructed by concatenating the remaining parts of the sentence that do not belong to the statement (see Pair 1 in Table \ref{table_cond_state_example}). In case the sentence has an exception (indicated by the phrase "provided, however"), we split it into two parts. The part before the phrase "provided, however" is processed as described above. The part after the phrase "provided, however" will be processed to extract a condition or statement, and will be negated (i.e., add the word "not" following \cite{jnlp_task5_coliee2021}) if necessary (see Pair 2 and 3 in Table \ref{table_cond_state_example}). For the query, we only consider the last sentence to extract the statement, and we consider other parts as the condition.

\textbf{Inference of YES/NO answer} To begin with, we employ TfIdf to select a condition from the extracted list of conditions that closely corresponds to the query's condition. After that, we evaluate the similarity between the retrieved condition and the query's condition, as well as the similarity between the corresponding statements. This is done by checking whether the number of negations in both the retrieved and the query's conditions and statements is odd or even. If both the condition and statement match, we answer YES, otherwise, the answer is NO.

% \usepackage{multirow}
\begin{table}[]
\resizebox{0.48\textwidth}{!}{%
\begin{tabular}{cl}
\multicolumn{1}{l}{} &   \\ \hline
\multicolumn{1}{|c|}{\textbf{Article}} & \multicolumn{1}{l|}{\begin{tabular}[c]{@{}l@{}}Article 23  \\ (1) If a person's domicile is unknown, the person's \\ residence is deemed to be the person's domicile.\\ (2) If a person does not have a domicile in Japan, the \\ person's residence is deemed to be the person's domicile, \\ regardless of whether the person is a Japanese national \\ or a foreign national; \textit{provided, however, that this does} \\ \textit{not apply if the law of domicile is to be applied in} \\ \textit{accordance with the provisions of the laws that establish} \\ \textit{the governing law.}\end{tabular}} \\ \hline
\multicolumn{1}{|c|}{\multirow{2}{*}{\textbf{Pair 1}}} & \multicolumn{1}{l|}{\textbf{Condition:} If a person's domicile is unknown}     \\ \cline{2-2} 
\multicolumn{1}{|c|}{}      & \multicolumn{1}{l|}{\begin{tabular}[c]{@{}l@{}}\textbf{Statement:} the person's residence is deemed to be the \\ person 's domicile\end{tabular}}      \\ \hline
\multicolumn{1}{|c|}{\multirow{2}{*}{\textbf{Pair 2}}} & \multicolumn{1}{l|}{\begin{tabular}[c]{@{}l@{}}\textbf{Condition:} If a person does not have a domicile in Japan \\ and regardless of whether the person is a Japanese \\ national or a foreign national\end{tabular}}  \\ \cline{2-2} 
\multicolumn{1}{|c|}{}      & \multicolumn{1}{l|}{\begin{tabular}[c]{@{}l@{}}\textbf{Statement:} the person's residence is deemed to be the \\ person 's domicile\end{tabular}}      \\ \hline
\multicolumn{1}{|c|}{\multirow{2}{*}{\textbf{Pair 3}}} & \multicolumn{1}{l|}{\begin{tabular}[c]{@{}l@{}}\textbf{Condition:} If a person does not have a domicile in Japan \\ and regardless of whether the person is a Japanese \\ national or a foreign national and if the law of domicile \\ is to be applied in accordance with the provisions of the \\ laws that establish the governing law\end{tabular}}      \\ \cline{2-2} 
\multicolumn{1}{|c|}{}      & \multicolumn{1}{l|}{\begin{tabular}[c]{@{}l@{}}\textbf{Statement:} the person's residence is \textbf{not} deemed to be \\ the person 's domicile\end{tabular}}  \\ \hline
\end{tabular}
}
\caption{\label{table_cond_state_example}Example of (condition, statement) pairs extracted from a legal article. Pair 1 is extracted from the sentence in point (1), pairs 2 and 3 are extracted from the sentence in point (2).}
\end{table}

\subsubsection{Third approach: SVM ensembles with condition-statement extraction}

This method involves using SVM on the data to generate a series of YES/NO predictions, which are then combined with predictions from the condition-statement extraction approach. To achieve this, we adopt an ensemble method that involves identifying whether a query is a specific-scenario query or a general query, as described in \cite{nguyen2022legal}. A specific-scenario query, as opposed to a general query, is a type of query that describes a particular situation or scenario in real life, requiring the ability to identify relevant abstract articles based on understanding the scenario. Specific-scenario queries are detected by identifying uppercase characters used to refer to legal persons or objects in the queries, following a consistent template observed in legal bar queries. 

We have found that the condition-statement extraction method tends to perform well with general queries, while SVM is better suited for specific-scenario queries. Consequently, when the two approaches produce different answers for a given query, we determine the final answer based on the query type: if it is a specific-scenario query, the final answer follows SVM's prediction, while for general queries, the final answer follows the condition-statement extraction method's prediction.

\section{Experiments}
In this section, we describe in detail the result of our methods introduced in Section~\ref{sec:method} in the development process and the final result from the COLIEE committee. 
\subsection{Task 2. Case Law Processing}
\subsubsection{Dataset and baselines} The statistics of the dataset provided by COLIEE competition are shown in Table \ref{table-41-1}. We compare our approach against the following baselines:
\begin{itemize} 
    \item BM25: We use the BM25 Lucene implementation in the Anserini\footnote{\url{https://github.com/castorini/pyserini}} open-source toolkit with the default parameters setting. We select the top \emph{k} candidates as the prediction with \emph{k} selected by grid search on the validation set.
    \item PT MonoT5-[large/3B]: We use the pre-trained MonoT5-large\footnote{castorini/monot5-large-msmarco} and MonoT5-3B \footnote{castorini/monot5-3b-msmarco} checkpoints as zero-shot baselines.
    \item PT BERT-large: We use a pre-trained BERT-large re-ranker\footnote{Luyu/bert-base-mdoc-bm25} on MS MACRO as an alternative approach to MonoT5. We use the implementation and the checkpoints released by the authors.
\end{itemize}
\begin{table}[H]
\begin{tabular}{ |c|c|c|c| } 
\hline
 & Train & Validation & Test \\
\hline
Size & 525 & 100 & 100 \\ 
Case ID range & 001 - 525 & 526 - 625 & 626 - 725 \\
Candidates / Case Ratio & 35.31 & 32.5 & 37.4 \\ 
Entailments / Case Ratio & 1.17 & 1.18 & 1.2 \\ 
\hline
\end{tabular}
\caption{\label{table-41-1}Dataset statistics}
\end{table}

\subsubsection{Submissions}
Our submissions are described below:
\begin{itemize} 
    \item FT MonoT5-large RS (Run name: \texttt{mt5l-e2}): The fine-tuned MonoT5-large using random negative sampling strategy.
    \item FT MonoT5-large HS (Run name: \texttt{mt5l-ed}): The fine-tuned MonoT5-large using our hard negative sampling strategy.
    \item Ensemble (Run name: \texttt{mt5l-ed4}): The ensembled top 5 checkpoints using the procedure in 3.1.2.
\end{itemize}
Except for BM25, we generate the prediction for all the baselines and submissions using the procedure in 3.1.3.

\subsubsection{Experimental results}
Table \ref{table:task2_vs_baseline} shows the validation F1 score of our methods compared with the baselines. Table \ref{table:task2_results} shows the results of the teams participating in the competition. It can be seen that our methods for this task outperform others by a considerable gap in the F1 score, indicating the appropriateness of our formulation for Task 2.

\begin{table}[H]
\begin{tabular}{ |l|c|c| } 
\hline
 \textbf{Method} & \textbf{Validation F1} \\
\hline
\textit{Baselines} & \\
+ BM25 & 61.47 \\ 
+ PT MonoT5-large & 68.62 \\ 
+ PT MonoT5-3B & 68.31 \\ 
+ PT BERT-large & 53.21 \\ 
% \hline
% + 3th Place & - & 71.82 \\
% + 5th Place & - & 69.30 \\
% + 6th Place & - & 68.18 \\
\hline
\textit{Our methods} & \\
+ FT MonoT5-large RS (\texttt{mt5l-e2}) & 75.23 \\ 
+ FT MonoT5-large HS (\texttt{mt5l-ed}) & 79.29 \\ 
+ Ensemble (\texttt{mt5l-ed4}) & \textbf{80.18} \\ 
\hline
\end{tabular}
\caption{\label{table:task2_vs_baseline}Our methods in Task 2 compared with baselines}
\end{table}

\begin{table}[h]
\centering
\begin{tabular}{|l|c|c|c|}
\hline
\textbf{Run} & \textbf{F1 (\%)} & \textbf{Precision (\%)} & \textbf{Recall (\%)} \\
\hline
\textbf{CAPTAIN.mt5l-ed} & \textbf{74.56} & 78.70 & \textbf{70.83} \\
\textbf{CAPTAIN.mt5l-ed4} & 72.65 & 78.64 & 67.50 \\
THUIR.thuir-monot5 & 71.82 & \textbf{79.00} & 65.83 \\
\textbf{CAPTAIN.mt5l-e2} & 70.54 & 75.96 & 65.83 \\
THUIR.thuir-ensemble\_2 & 69.30 & 73.15 & 65.83 \\
JNLP.bm\_cl\_1\_pr\_1 & 68.18 & 75.00 & 62.50 \\
IITDLI.iitdli\_task2\_run2 & 67.27 & 74.00 & 61.67 \\
JNLP.cl\_1\_pr\_1 & 65.45 & 72.00 & 60.00 \\
UONLP.test\_no\_labels & 63.87 & 64.41 & 63.33 \\
THUIR.thuir-ensemble & 60.91 & 67.00 & 55.83 \\
NOWJ.non-empty & 60.79 & 64.49 & 57.50 \\
NOWJ.hp & 60.36 & 65.69 & 55.83 \\
IITDLI.iitdli\_task2\_run3 & 53.04 & 55.45 & 50.83 \\
LLNTU.task2\_llntukwnic & 18.18 & 20.00 & 16.67 \\
\hline
\end{tabular}
\caption{\label{table:task2_results}Results of Task 2 in the competition}
\end{table}

By looking at the F1 results, we can see that the fine-tuned MonoT5 on the legal entailment dataset outperforms the pre-trained MonoT5 on MS MARCO \cite{nguyen2016ms} by a significant margin. The introduced hard negative mining procedure proved to be a more effective method for fine-tuning than the vanilla random sampling strategy. The ensemble approach has the best performance on the validation set but is inferior to a single model fine-tuned MonoT5 on the test set, which is likely the result of overfitting when performing hyperparameter grid searching. And finally, our results show that the sequence-to-sequence approach has better performance in textual entailment tasks compared to the BERT re-ranker approach.

\subsection{Task 3. The Statute Law Retrieval Task}
\subsubsection{Dataset} Similar to the previous year's format, the dataset of this task consists of 996 questions, a legal corpus (Civil Code) with 768 articles, and 1272 pairs of questions and relevant articles (positive samples).  The examples of this dataset are shown in Table~\ref{table_cond_state_example}. For the development process, we choose questions that have an ID starting with \texttt{R02} (81 questions) or \texttt{R03} (109 questions) as a validation set and conduct model/settings evaluations on this sub-set.

\subsubsection{Submissions}

We describe three settings we submitted in the COLIEE Task-3 submission time. 

\begin{itemize}
    \item \texttt{CAPTAIN.bjpAll}: We ensemble all the sub-models from the pre-trained language model  \textit{cl-tohoku/bert-base-japanese-whole-word-masking}. This submission is the ensemble result of five checkpoints in \textit{MCkpt} setting,  five checkpoints in \textit{DatFlt-q} setting, and the best checkpoint in \textit{DatFlt-a} setting. 
    \item \texttt{CAPTAIN.allEnssMissq}: We ensemble the sub-models from the pre-trained language model  \textit{cl-tohoku/bert-base-japanese-whole-word-masking} and \textit{monot5-large-msmarco-10k}. This submission is the ensemble result of the best checkpoint in \textit{MCkpt} setting and \textit{monot5-large-msmarco-10k} fine-tuned model. 
    \item \texttt{CAPTAIN.allEnssBoostraping}:  We ensemble  the sub-models from the pre-trained language model  \textit{cl-tohoku/bert-base-japanese-whole-word-masking} and \textit{monot5-large-msmarco-10k}. This submission is the ensemble result of the best checkpoint in \textit{MCkpt-a} setting and \textit{monot5-large-msmarco-10k} fine-tuned model. 
    
\end{itemize}

\subsubsection{Experiments and Results}

We conducted experiments to estimate the effectiveness of our proposed methods based on the result of the development set and compared it with the best result \cite{yoshioka2023hukb} of COLIEE 2022.  Especially for a fair comparison with their result, we only optimize model hyper-parameters based on the \texttt{R02} sub-set and evaluate the result on \texttt{R03} sub-set. The best models found in the development process are used to predict the output of the official test set \texttt{R04}.   

\paragraph{Hyper-parameters} In the fine-tuning process, we used five epochs, the learning rate is selected in $\{1e^{-5}, 2e^{-5}\}$, the maximum token length is 512 sub-word tokens and batch size is selected in $\{16, 32\}$. The top 5 best model checkpoints are saved based on the macro F2 scores of development set \texttt{R02}.

\paragraph{Development results}  Our main development results are shown in Table~\ref{tab:taske3_devresults}. 
Firstly, we develop our system based on two main pre-trained language models: monoT5 versions \textit{castorini/monot5-large-msmarco }\footnote{\url{https://huggingface.co/castorini/monot5-large-msmarco}},  \textit{castorini/monot5-large-msmarco-10k}\footnote{\url{https://huggingface.co/castorini/monot5-large-msmarco-10k}} and  Japanese pre-trained language model \textit{cl-tohoku/bert-base-japanese-whole-word-masking}\footnote{\url{https://huggingface.co/cl-tohoku/bert-base-japanese-whole-word-masking}}:
\begin{itemize}
    \item \textit{monoT5 pre-trained model}: This pre-trained model is specialized for text retrieval tasks trained on the MACRO large-scaled dataset. Based on the suggestion of the authors \cite{nogueira-etal-2020-document}, we used version \textit{castorini/monot5-large-msmarco-10k} for zero-shot learning and ensemble with version \textit{castorini/ monot5-large-msmarco} fine-tuned on the COLIEE dataset. The result of the development set proved that incorporating two versions boosts the performance by about 2\% F2 score (Table~\ref{tab:taske3_devresults}). Therefore, we used the ensemble result of this model as the main prediction when incorporated with other settings.  
    \item \textit{Japanese pre-trained model}: We mainly evaluate our proposed method on a pre-trained language model (\textit{cl-tohoku/bert-base-japanese-whole-word-masking}) because this model did not learn any text retrieval task. This makes it straightforward to measure the effectiveness of our method on legal text retrieval tasks (two last rows in Table~\ref{tab:taske3_devresults}). In the comparison between three settings \textit{DatFlt-q}, \textit{DatFlt-a}, \textit{MCkpt} separately, we found that \textit{DatFlt-q} setting shows the strength in the Recall metric but low on the Precision. Since this setting is strengthened in \textit{"missed"} questions, which makes the model reduces the number of \textit{missed} questions as much as possible. In contrast to setting \textit{DatFlt-a}, which prefers to return the accurate articles that make the Precision typically higher than \textit{DatFlt-q}.  
    \item \textit{Ensemble model}: We combine the result of monoT5 with Japanese pre-trained models by the \textit{main-auxiliary ensemble method} (described in section~\ref{subs:3}) with monoT5 play as the main model.  Besides, we also combine all the sub-models of our proposed methods for comparison (last rows in Table~\ref{tab:taske3_devresults}).  These results show that the effective incorporation between our methods and monoT5 achieved significant improvement in the overall system. 
\end{itemize}

\begin{table}[h]
\centering
\resizebox{0.48\textwidth}{!}{%
\begin{tabular}{|l|c|c|c|}
\hline
\textbf{Run} & \textbf{F2 (\%)} & \textbf{P (\%)} & \textbf{R (\%)} \\
\hline
\multicolumn{4}{c}{Evaluate on R02 questions} \\ \hline
monoT5-L-10k (zs) & 68.29 & 68.93 & 69.14 \\
monoT5-L (ft)   & 68.29 & 68.93 & 69.14 \\
monoT5-L (ft)  + monoT5-L-10k (zs) & 70.33 & 71.81 & 70.99 \\
\hdashline
DatFlt-q & 69.35 & 53.81 &    \textbf{80.86}  \\
DatFlt-a & 69.08 & 58.09&  77.78 \\ 
MCkpt & 68.38 &  61.21  &  72.84 \\
\hdashline
DatFlt-q + \underline{monoT5} (\texttt{allEnssMissq}) &  \textbf{76.36} &   \textbf{74.28} & 78.40\\ 
DatFlt-a + \underline{monoT5} (\texttt{allEnssBoostraping}) &  {76.29} &  73.66  & 78.40 \\ 
DatFlt-q + DatFlt-a + \underline{MCkpt}   (\texttt{bjpAll}) &  70.45 & 63.37  & 75.31\\
 \hline
\multicolumn{4}{c}{Evaluate on R03 questions} \\ 
\hline
HUKB2 \cite{yoshioka2023hukb} & 82.04 & 81.80 & 84.05 
\\
\hdashline
monoT5-L-10k  (zs) & 79.13 & 82.19 & 79.62 \\
monoT5-L (ft)  & 80.05 & 83.10 & 80.54 \\
monoT5-L (ft)  + monoT5-L-10k (zs) & 81.28 & 81.61 & 83.29 \\
\hdashline
DatFlt-q & 75.84 & 59.06 &  \textbf{88.84}  \\ 
DatFlt-a & 76.90& 62.58 & 87.92  \\ 
MCkpt & 80.68 &  76.07  & 85.06 \\
\hdashline
DatFlt-q + \underline{monoT5} (\texttt{allEnssMissq})  & \textbf{85.52} & 84.56 & 87.45\\ 
DatFlt-a + \underline{monoT5} (\texttt{allEnssBoostraping}) & 85.01 &   \textbf{85.47}  & 86.35\\ 
DatFlt-q + DatFlt-a + \underline{MCkpt}  (\texttt{bjpAll})  &  82.15 & 75.84  & 87.81 \\\hline
\end{tabular}
}
\caption{\label{tab:taske3_devresults}Results of Task 3 on the development set. The notations (ft), (zs) refer to \textit{fine-tuned} on the COLIEE dataset and \textit{zero-shot} learning, respectively.  }
\end{table}

\paragraph{Official test results} Table~\ref{table:task3_testresults} shows the full test results of Task 3 provided by the COLIEE organization. Our results got first and second places and also competitive in third place from other teams.  The results demonstrate that our proposed methodology exhibits a high degree of generalizability and holds significant promise for applications in text retrieval tasks. Besides, we also got the best performance on most other metrics, such as MAP, R@5, R@10, and R@30, which is clear evidence of the effectiveness of our method. In addition, in the two settings \texttt{allEnssMissq} and \texttt{allEnssBoostraping}, because we combine two different pre-trained models with monoT5 model playing as a primary model for prediction, the evaluation scores on four metrics (MAP, R@5, R@10, and R@30) are exactly same. In the  submission \texttt{bjpAll}, (equal to \textit{DatFlt-q + DatFlt-a + \underline{MCkpt}}), the ranking scores of our proposed method still work effectively because all the models of these settings are the different checkpoints of the same pre-trained model in the fine-tuning process, which share the same semantic space. 

\begin{table}[h]
 \centering
 \resizebox{0.48\textwidth}{!}{%

\begin{tabular}{|l|c|c|c|c|c|c|c|}
\hline
\textbf{Run} & \textbf{F2} & \textbf{P} & \textbf{R} & \textbf{MAP} & \textbf{R5} & \textbf{R10 } & \textbf{R30} \\
\hline
\textbf{CAPTAIN.Missq} & \textbf{75.69} & \textbf{72.61} & 79.21 & 69.21 & 75.38 & 83.85 & 88.46 \\
\textbf{CAPTAIN.Boost} & 74.70 & 71.62 & 78.22 & 69.21 & 75.38 & 83.85 & 88.46 \\
JNLP3 & 74.51 & 64.52 & \textbf{82.18} & 70.99 & 80.00 & 83.85 & 90.00 \\
\textbf{CAPTAIN.bjpAll} & 74.15 & 70.63 & 77.72 & \textbf{84.64} & \textbf{87.69} & \textbf{90.77} & \textbf{96.15} \\
NOWJ.ensemble & 72.73 & 68.23 & 76.73 & 78.99 & 78.46 & 80.77 & 89.23 \\
HUKB1 & 67.25 & 62.79 & 70.79 & 73.97 & 75.38 & 83.85 & 93.08 \\
JNLP2 & 66.28 & 64.22 & 70.30 & 68.61 & 73.08 & 79.23 & 88.46 \\
HUKB3 & 66.25 & 65.02 & 68.32 & 74.14 & 74.62 & 84.62 & 93.08 \\
JNLP1 & 65.71 & 66.50 & 67.82 & 68.65 & 73.08 & 79.23 & 88.46 \\
LLNTUgigo & 65.35 & 73.27 & 64.36 & 76.43 & 80.00 & 88.46 & 91.54 \\
HUKB2 & 64.85 & 67.82 & 65.84 & 74.14 & 74.62 & 84.62 & 93.08 \\
LLNTUkiwiord & 63.26 & 70.30 & 62.38 & 76.25 & 82.31 & 86.92 & 93.08 \\
UA.TfIdf\_threshold2 & 56.42 & 62.05 & 56.44 & 65.51 & 66.92 & 79.23 & 84.62 \\
UA.TfIdf\_threshold1 & 55.45 & 63.37 & 54.46 & 65.51 & 66.92 & 79.23 & 84.62 \\
UA.BM25 & 55.01 & 63.37 & 53.96 & 64.86 & 66.92 & 79.23 & 86.92 \\
NOWJ.ttlJP & 06.37 & 05.78 & 06.93 & 07.16 & 06.15 & 07.69 & 13.08 \\
NOWJ.COENJP & 01.82 & 01.49 & 01.98 & 02.46 & 01.54 & 03.85 & 06.15 \\
\hline
\end{tabular}
}
\caption{\label{table:task3_testresults}Results on the official test of Task 3. "CAPTAIN.Missq" stands for CAPTAIN.\texttt{allEnssMissq}, "CAPTAIN.Boost" stands for CAPTAIN.\texttt{allEnssBoostraping}.}
\end{table}

\subsection{Task 4. The Legal Textual Entailment Task}
\subsubsection{Dataset} The dataset we use for experiments in Task 4 is similar to the dataset used for Task 3, but for the task of textual entailment. It contains 996 questions, and a legal corpus (Civil Code) with 768 articles. For the validating and testing step, we used the questions in R01 (111 questions), R02 (81 questions), and R03 (106 questions). The remaining questions are used for the training process.   

\subsubsection{Submissions}
In Task 4, we submit 3 runs as follows:
\begin{itemize}
\item CAPTAIN.gen: The predictions of the first approach (\textit{online data augmentation})
\item CAPTAIN.run1: The predictions of the second approach (condition-statement extraction)
\item CAPTAIN.run2: The predictions of the third approach (SVM ensembles with condition-statement extraction)
\end{itemize}

\subsubsection{Experiments and results}
For the first approach experiments, we use the pre-trained model \textit{cl-tohoku/bert-base-japanese-whole-word-masking} to re-sample the training data. The training data, in this case, involves remaining data and augmented data generated randomly by \textit{online data augmentation} mechanism. For the second and the third approaches, we only use the original data.

The results of our approaches for Task 4 with the validation data are shown in Table \ref{task4_result}. The results of Task 4 in the competition are shown in Table \ref{table_results_task_4}.

\begin{table}[H]
    \centering
    \begin{tabular}{|l|c|c|c|}
    \hline
        Method & R01  & R02 & R03\\
    \hline
        CAPTAIN.run1 & 60.36 & 50.62 & 64.22\\
        CAPTAIN.run2 & 60.36 & 51.85 & 66.05\\
        CAPTAIN.gen & 58.56 & 67.90 & 56.88\\    \hline
    \end{tabular}
    \caption{Performance (accuracy) of Task 4 validation sets: R01, R02, R03}
    \label{task4_result}
\end{table}

\begin{table}[ht]
\centering
\begin{tabular}{|l|c|}
\hline
\textbf{Run} & \textbf{Accuracy (\%)} \\
\hline
JNLP3 & \textbf{78.22} \\
TRLABS\_D & 78.22 \\
KIS2 & 69.31 \\
UA-V2 & 66.34 \\
AMHR01 & 65.35 \\
LLNTUdulcsL & 62.38 \\
HUKB2 & 59.41 \\
\textbf{CAPTAIN.gen} & 58.42 \\
\textbf{CAPTAIN.run1} & 57.43 \\
NOWJ.multi-v1-jp & 54.46 \\
\textbf{CAPTAIN.run2} & 52.48 \\
NOWJ.multijp & 52.48 \\
NOWJ.multi-v1-en & 48.51 \\
\hline
\end{tabular}
\caption{\label{table_results_task_4}Results of Task 4 in the competition}
\end{table}

Through the results, we can see that the approach using the \textit{online data augmentation} mechanism has stable results on all testing and validating datasets. This shows that the data generated through \textit{online data augmentation} mechanism provided the model useful information to decide the final result. On the other side, extracting condition, and statement pairs from articles also gives positive results on the validation dataset. The performance of the \textit{condition and statement extracting} approach is close to the performance of \textit{online data augmentation}. However, when combined with the SVM model, the performance of the \textit{condition and statement extracting} approach is decreased as compared to the results in the validating dataset. The SVM model seems to be biased with the validation dataset, since that causes the reduction of performance on official testing questions. 

\section{Conclusions}
This paper describes our methods for addressing Task 2, Task 3, and Task 4 in COLIEE 2023. We have utilized appropriate deep learning techniques and applied rigorous engineering practices and methodologies to the competition, resulting in exceptional performance in these tasks. Going forward, we plan to further explore the properties of legal documents and queries to gain deeper insights and develop more effective techniques.

\begin{acks}
This work is partly supported by AOARD grant FA23862214039.
We express our gratitude to the reviewers for their valuable and constructive feedback.
\end{acks}

\bibliographystyle{ACM-Reference-Format}
\bibliography{references}

\appendix

\end{document}